\documentclass{article}
\usepackage{spconf,amsmath,epsfig}
\usepackage{booktabs}
\usepackage{multirow}
\usepackage{siunitx}
\usepackage{hyperref}

\title{SAR Despeckling using Overcomplete Convolutional Networks} 
\name{Malsha V. Perera, Wele Gedara Chaminda Bandara, Jeya Maria Jose Valanarasu, and Vishal M. Patel\thanks{\copyright \; 2022 IEEE. Personal use of this material is permitted. Permission from IEEE must be obtained for all other uses, in any current or future media, including reprinting/republishing this material for advertising or promotional purposes, creating new collective works, for resale or redistribution to servers or lists, or reuse of any copyrighted component of this work in other works.}}
\address{Johns Hopkins University\\
Department of Electrical and Computer Engineering\\
\{jperera4, wbandar1, jvalana1, vpatel36\}@jhu.edu}
\begin{document}
%
\maketitle
\begin{abstract}
Synthetic Aperture Radar (SAR) despeckling is an important problem in remote sensing as speckle degrades SAR images, affecting downstream tasks like detection and segmentation. Recent studies show that convolutional neural networks (CNNs) outperform classical despeckling methods. Traditional CNNs try to increase the receptive field size as the network goes deeper, thus extracting global features. However, speckle is relatively small, and increasing receptive field does not help in extracting speckle features. This study employs an overcomplete CNN architecture to focus on learning low-level features by restricting the receptive field. The proposed network consists of an overcomplete branch to focus on the local structures and an undercomplete branch that focuses on the global structures. We show that the proposed network improves despeckling performance compared to recent despeckling methods on synthetic and real SAR images. Our code is available at : \url{https://github.com/malshaV/sar_overcomplete}
\end{abstract}

\section{Introduction}

\setlength{\belowdisplayskip}{0pt} \setlength{\belowdisplayshortskip}{0pt}
\setlength{\abovedisplayskip}{0pt} \setlength{\abovedisplayshortskip}{0pt}

\label{sec:intro}
Synthetic Aperture Radar (SAR) is a coherent imaging modality and has the advantage of being able to operate  all-day and in all-weather conditions. Therefore, SAR images are an important complementary source of information with respect to optical images. 
However, SAR images are often affected by multiplicative speckle noise. The presence of speckle can adversely affect the interpretability and  processing of  SAR images. Hence, the removal of speckle from SAR images is crucial for improving the performance of high-level tasks.

Given a SAR intensity image $Y$ and speckle-free SAR image $X$, the mathematical model of SAR can be expressed as follows:
\begin{equation}
   Y = XN
\label{eq1},
\end{equation}
where $N$ is the multiplicative speckle.  Under the hypothesis of fully developed speckle, $N$ follows a Gamma distribution:
\begin{equation}
    p(N) = \frac{1}{\Gamma(N)}L^LN^{L-1}e^{-LN},
\label{eq2}
\end{equation}
where $\Gamma(.)$ is the Gamma function and $L$ is the number of looks of the SAR image.

Traditionally, filter-based approaches have been widely  used for SAR despeckling. Lee filter \cite{lee_filter} and Kuan filter \cite{kuan_filter} are examples of despeckling algorithms which use a local filter-based approach. These methods use the spatial correlation of image pixels to filter coherent noise with the use of a sliding window. More recently, non-local filtering based algorithms such as PPB \cite{ppb} and SAR-BM3D \cite{sar_bm3d} have  been proposed for SAR image despeckling. 

In recent years, deep learning algorithms have gained popularity and achieved state-of-the-art performance in many computer vision tasks. Following this, a few recent studies have attempted to apply deep networks for SAR despeckling. Chierchia \textit{et al.} \cite{SARCNN} proposed SAR-CNN, which performs despeckling by applying a Convolutional Neural Network (CNN) for despeckling. SAR-CNN transforms the SAR images to the homomorphic form to convert the multiplicative noise to an additive noise. They obtain clean data for training using multi-temporal fusion and use a loss function based on similarity measure for speckle noise distribution. ID-CNN \cite{IDCNN} proposed by Wang \textit{et al.} uses a residual architecture which aims at estimating speckle from the original-domain image and the despeckled image is obtained by dividing the input image by the estimated speckle. Unlike SAR-CNN, this study uses synthetically speckled optical images to train the network. In \cite{KLD}, ID-CNN network is trained with composite loss function comprising of a Mean Square Error (MSE) and Kullback-Leibler divergence (KL) between the predicted and simulated speckle probability distribution. Liu \textit{et al.} \cite{MRDDANet} proposed a multiscale residual dense dual attention network (MRDDANet) for despeckling which focuses on suppressing the speckle while fully retaining the texture details of the image. 

Most deep learning-based SAR image despeckling methods use either an ``encoder-decoder'' architecture or a residual architecture. The receptive field in the latent space of these architectures are quite large. Hence, they focus on extracting global features than local features. While extracting global features is important for many computer vision tasks, applications like SAR image despeckling require better local feature extraction as the speckle is small and can be better detected with the use of small receptive fields. Overcomplete convolution architectures have the ability to extract low-level features of an image by restricting the receptive field from enlarging in deeper layers. The overcomplete architecture \cite{OUCD} achieves this by transforming the image into a representation with higher resolution by replacing the max-pooling layers in an undercomplete CNN architecture with upsampling layers.

To this end, in this paper, we propose an overcomplete convolutional architecture \cite{OUCD} to extract meaningful low-level features which captures and removes speckle effectively. We also combine an undercomplete architecture to our proposed network, in order to allow the network to learn high level features which help in the reconstruction of large structures in a given image. This way we combine both the overcomplete and undercomplete architectures to perform despeckling effectively. We train our proposed network using synthetically speckled optical images and test our network on both synthetic and real SAR images.

\section{Proposed Method}
\label{sec:method}

\begin{figure*}[htbp]
\centerline{\includegraphics[width=.85\textwidth]{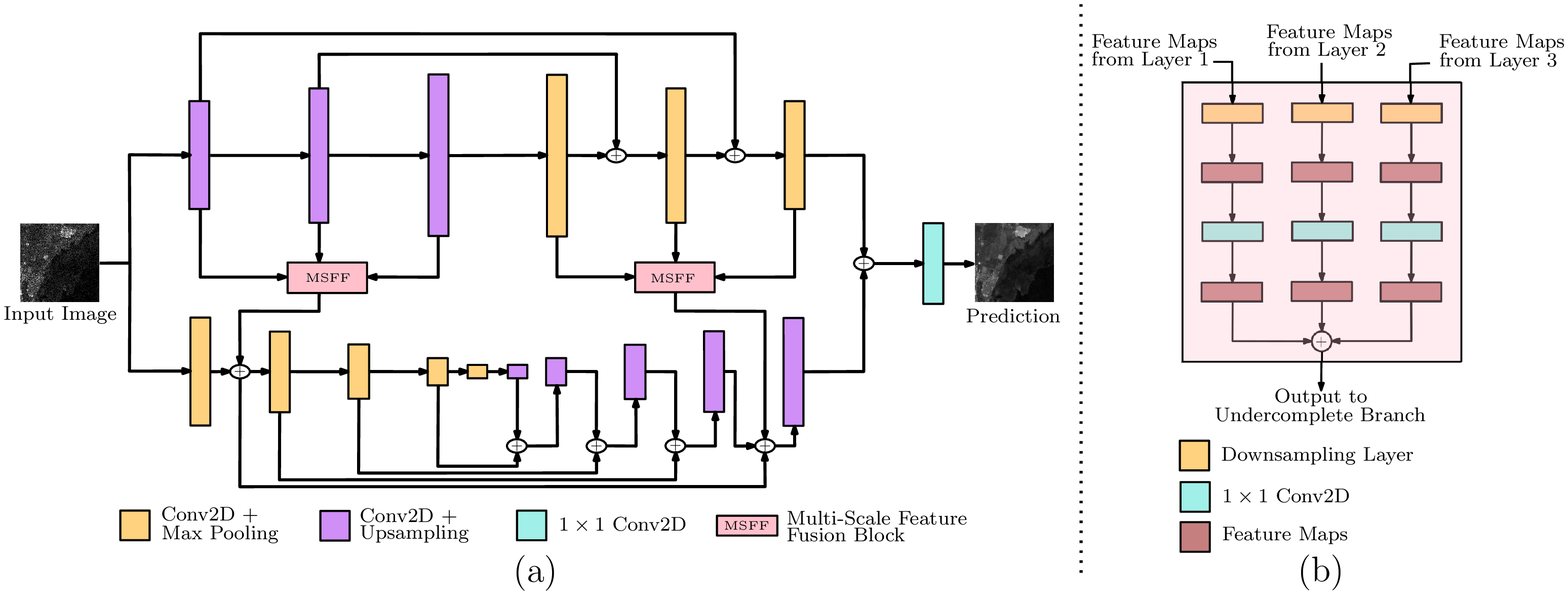}}
\vskip-10pt\caption{(a) Overview of the proposed network architecture (b) Multi-Scale Feature Fusion Block .}
\label{network}
\end{figure*}

\noindent {\bf{Network architecture.}} 
Fig. \ref{network} illustrates an overview of the proposed network. The proposed network consists of two branches: overcomplete and undercomplete branch. The overcomplete branch comprises of an upsampling encoder pathway and a downsampling decoder pathway, while the undercomplete branch consists of a downsampling encoder pathway and an upsampling decoder pathway. The features of the overcomplete branch is transferred to the undercomplete branch with the use of Multi-Scale Feature Fusion (MSFF) block.

The overcomplete branch which focuses more on capturing  local features has $3$ convolutional blocks in both encoder and decoder. The convolutional block in the encoder consists of a $3\times3$ convolution layer followed by an upsampling layer and a ReLU activation. The upsampling layer performs bilinear upsampling to its input with a factor of $2$. The convolutional blocks in the decoder of the overcomplete branch consist of a $3\times3$ convolution layer followed by a $2\times2$ maxpooling layer of stride 2 and a ReLU activation. Furthermore, we add skip connections between the encoder and the decoder of the overcomplete branch as shown in Fig. \ref{network}. By adding these skip connections, we ensure that the network achieves better localization. 

The undercomplete branch allows the network to learn global features in the given image. Therefore, by adding an undercomplete branch, we allow the despeckling network to incorporate global features as they still have some meaningful information helpful for despeckling. This branch of the proposed network closely resembles the standard U-Net architecture.  Both the encoder and the decoder of the undercomplete branch consist of $5$ convolutional blocks. The convolutional blocks in the encoder consist of a $3 \times 3$ convolutional layer followed by a $2 \times 2$ maxpooling layer of stride $2$ and a ReLU activation layer. Each convolutional block in the decoder has a $3 \times 3$ convolutional layer, upsampling layer of scale factor 2 and a ReLU activation layer, in the given order. Similar to the standard U-Net architecture, the undercomplete branch also has skip connections connecting the encoder and decoder features. Finally, the output features from the overcomplte and undercomplete branches are added together before passing through a $1 \times 1$ convolutional layer to get the final prediction.\\ 
\noindent {\bf{Multi-Scale Feature Fusion Block.}} 
In the proposed network, the MSFF block \cite{OUCD} is used to transfer the low level features of the overcomplete branch with different scales to the undercomplete branch. Feature maps of all the layers in the overcomplete branch can capture finer details which is important when removing fine speckles. However, the layers of the undercomplete branch would not capture such fine details. By transferring the feature maps from the overcomplete branch to the undercomplete branch, we allow the undercomplete branch to learn better global features. To this end, we take the outputs of the first three convolutional blocks of the encoder of the overcomplete branch and pass them through an MSFF block to the output of the first convolution block of the undercomplete branch. The task of the MSFF block is to transform the outputs from the overcomplete branch into the similar scales as undercomplete branch while maintaining equal number of feature maps. Similarly, the output of the last three convolutional blocks of the overcomplete branch is passed through an MSFF block to the input of the last convolutional block of the undercomplete branch. The architecture of MSFF is illustrated in Fig. \ref{network} (b). First, each input to the MSFF  block is downsampled to match the scale of the feature maps in the
undercomplete branch where the final output of MSFF block is going to be added. The downsampling is performed using bilinear interpolation. The downsampled feature maps are passed through $1 \times 1$ convolutions in order to get equal number of feature maps across different scales of the overcomplete network. Finally, the three outputs from each $1 \times 1$ convolution are added together and passed to the undercomplete branch.\\
\noindent {\bf{Loss functions.}}
We propose a composite loss function to train the network end-to-end. The proposed  loss function comprises of $l_2$ loss ($\mathcal{L}_{l_2}$) and total variation loss ($\mathcal{L}_{TV}$). The $l_2$ loss is widely used in image restoration tasks. In order to encourage smoothness while preserving edges, we add $\mathcal{L}_{TV}$ loss to the composite loss function. Given the prediction $\hat{X}$ and ground truth $X$, the composite loss $\mathcal{L}$ is defined as follows:
\begin{equation}
    \mathcal{L}_{Total} = \mathcal{L}_{l_2} + \lambda\mathcal{L}_{TV},
\end{equation}
where
\begin{equation}
    \mathcal{L}_{l_2} = \| \hat{X} - X \|_2^2,
\end{equation}
\begin{equation}
    \mathcal{L}_{TV} = \sum_{i,j} |\hat{X}_{i+1,j} - \hat{X}_{i,j}| + | \hat{X}_{i,j+1} - \hat{X}_{i,j}|,
\end{equation}
and $\lambda$ is the weight assigned to the total variation loss.

\section{Experiments and Results}
\label{sec:experiments}

\begin{figure*}[ht]
\centerline{\includegraphics[width=\textwidth]{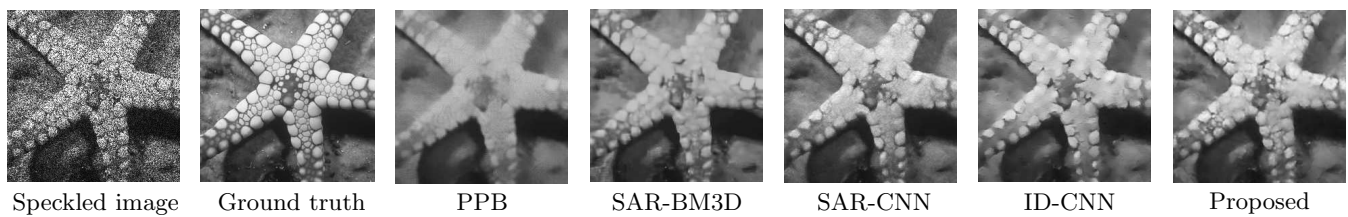}}
\vspace{-3mm}
\caption{Qualitative results of different despeckling methods on synthetic images.}
\label{syn_results_img}
\end{figure*}

\begin{figure*}[ht]
\centerline{\includegraphics[width=\textwidth]{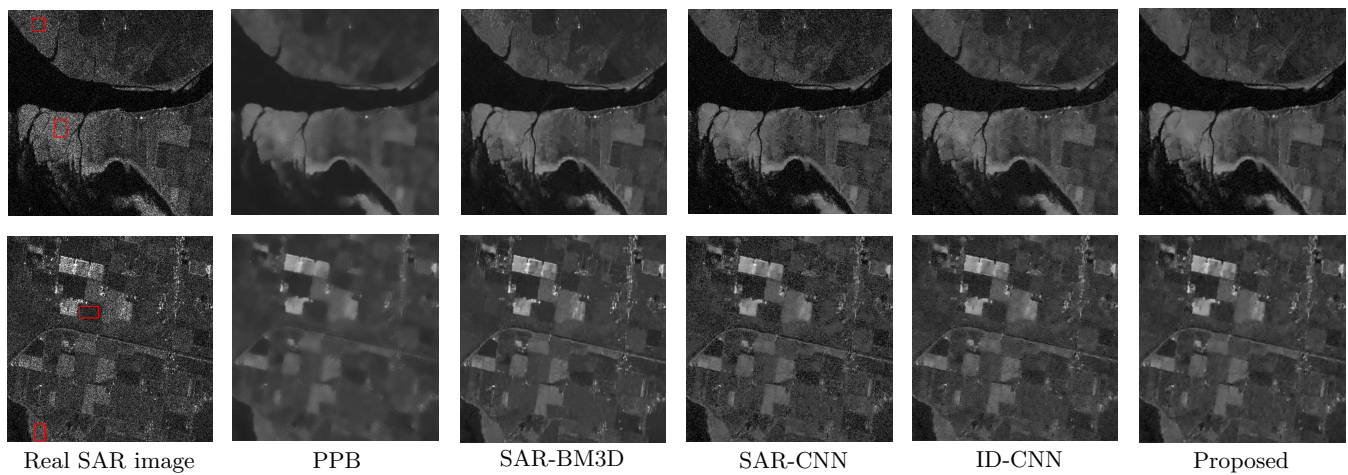}}
\vspace{-3mm}
\caption{Qualitative results of different despeckling methods on real SAR images.}
\label{real_results_img}
\end{figure*}
In this section, we present the experiments and results of our proposed method on both synthetic and real SAR images. We compare the performance of our method with the following traditional and CNN-based despeckling algorithms: PPB \cite{ppb}, SAR-BM3D \cite{sar_bm3d}, SAR-CNN \cite{SARCNN} and ID-CNN \cite{IDCNN}. In order to train the proposed network, we employ images from the  Berkeley segmentation dataset (BSD) \cite{BSD} to generate a synthetic SAR-like dataset. The single-look synthetic speckled images were created according to equations \ref{eq1} and \ref{eq2} by setting $L = 1$. We use 450 images from the BSD dataset for training and the rest are used for validation. To train the proposed network, we augment the speckled input image by randomly cropping $128 \times 128$ patches. We trained the proposed method using the ADAM optimizer with a learning rate of 0.0002 for 400 epochs, while setting $\lambda = 5 \times 10^{-5}$. The proposed network was implemented using PyTorch  and all experiments were performed using an NVIDIA RTX 2080Ti GPU. 

\begin{table}[h]
\setlength\tabcolsep{0pt}
\caption{Average results on synthetic images of Set12 dataset.}\label{results table}
\centering
\smallskip
\begin{tabular*}{\columnwidth}{@{\extracolsep{\fill}}lcc}
\toprule
Method & PSNR & SSIM  \\
 \midrule
  PPB \cite{ppb}& 21.90 & 0.599  \\
  SAR-BM3D \cite{sar_bm3d}& 23.51 & 0.701 \\
  SAR-CNN \cite{SARCNN}& 24.51 & 0.651\ \\
  ID-CNN \cite{IDCNN}& 24.44 & 0.685\\
  Proposed method & \textbf{24.89} & \textbf{0.722}\\
\bottomrule
\label{results_synthetic}
\end{tabular*}
\vspace{-7mm}
\end{table}

Table \ref{results_synthetic} shows the performance of the proposed method on synthetic speckled images generated using a set of well-known testing images \cite{testdata} in terms of Peak Signal-to-Noise ratio (PSNR) and Structured Similarity Index (SSIM). It can be noted from Table \ref{results table} that our proposed network outperforms popular traditional and CNN-based despeckling algorithms in terms of both PSNR and SSIM. Moreover, Fig. \ref{syn_results_img} shows results on a selected synthetic speckle image for visual comparison. It can be observed that by using an overcomplete network architecture, our proposed method performs better when restoring smaller structures in the speckled image compared to other CNN-based despeckling methods.

In Table \ref{results real table}, we present the results of our proposed method on real SAR images in terms of Equivalent Number of Looks (ENL) and the coefficient of variation (Cx). ENL is given by the ratio between the square of the mean and the variance of a homogeneous region. Cx is a measure of texture and is defined as the ratio between standard deviation and the mean intensity of a region. The regions used to calculate ENL and Cx are indicated in red boxes in Fig. \ref{real_results_img}. From the results, we can observe that our proposed method resulted in the best despeckling performance with the highest ENL values and lowest Cx values in each region. Fig. \ref{real_results_img} shows despeckled results on real SAR images and we can observe that the CNN-based methods tends to show a better speckle suppression than the traditional methods. However, SAR-CNN and ID-CNN tends to oversmooth the image when despeckling. With the use of overcomplete architecture, our proposed method is able to reduce oversmoothing and recover fine details when restoring the images.

\begin{table}[h]
\setlength\tabcolsep{0pt}
\caption{Results on real SAR images.} 
\label{results real table}
\centering
\smallskip
\begin{tabular*}{\columnwidth}{@{\extracolsep{\fill}}lSSSSSSSS}
    \toprule
    \multirow{2}{*}{Method} &
      \multicolumn{2}{c}{Region 1} &
      \multicolumn{2}{c}{Region 2} &
      \multicolumn{2}{c}{Region 3} &
      \multicolumn{2}{c}{Region 4}\\
      & {ENL$\uparrow$} & {Cx$\downarrow$} & {ENL$\uparrow$} & {Cx$\downarrow$} & {ENL$\uparrow$} & {Cx$\downarrow$} & {ENL$\uparrow$} & {Cx$\downarrow$}\\
      \midrule
    PPB  & \text{87.0} & \text{0.11}  & \text{125.5} & \text{0.09} & \text{21.0} & \text{0.22} & \text{117.8} & \text{0.09} \\
    SARBM3D&  \text{110.8} & \text{0.09} & \text{104.2} & \text{0.10} & \text{34.9} & \text{0.17} & \text{122.9}  &\text{0.09}\\
    SARCNN & \text{87.4} & \text{0.11} & \text{51.12} & \text{0.14} & \text{30.7} & \text{0.18} & \text{68.6}& \text{0.12}\\
    IDCNN & \text{47.6} & \text{0.14} & \text{33.3} & \text{0.17} & \text{23.1} & \text{0.21} & \text{34.5} & \text{0.17}\\
    Proposed & \textbf{137.2} & \textbf{0.09} & \textbf{171.5} & \textbf{0.08} & \textbf{38.8} & \textbf{0.16} & \textbf{139.4} & \textbf{0.08}\\
    
    \bottomrule
  \end{tabular*}
  \vspace{-5mm}
\end{table}

\section{CONCLUSION}
\label{sec:conclusion}
We proposed a CNN with  an overcomplete and an undercomplete branch for despeckling. Using an overcomplete architecture allows the network to capture low level features and finer details than generic convolutional neural networks. The undercomplete branch of our proposed method makes sure that the network is able to capture global features as well. Experiments on synthetic and real SAR images show that the proposed method improves the  despeckling performance over popular traditional and CNN-based approaches while preserving fine details. 

\vspace{-3mm}

\section{ACKNOWLEDGEMENT}
This work was supported by the NSF CAREER Award under Grant 2045489.

\bibliographystyle{IEEEbib}
{\footnotesize \bibliography{refs}}

\end{document}